\documentclass[10pt,twocolumn,letterpaper]{article}
\usepackage{3dv}

\usepackage[utf8]{inputenc}
\usepackage[T1]{fontenc}
\usepackage{booktabs}
\usepackage{amsfonts}
\usepackage{nicefrac}
\usepackage{microtype}
\usepackage{overpic}
\usepackage{xcolor}
\usepackage{verbatim}

\usepackage{epsfig}
\usepackage{graphicx}
\usepackage{times,helvet,courier}
\usepackage[tight,footnotesize]{subfigure}
\usepackage{amsmath,amssymb,amsthm,bm}
\usepackage{multirow,makecell,footmisc,url}
\usepackage[ruled]{algorithm2e}
\usepackage{color}

\usepackage{hyperref}

\def \saliency {\textup{\saliency}}

\def \path {\mathit{path}}

\def \label {\mathit{label}}

\let\bs=\textbf

\threedvfinalcopy

\def\threedvPaperID{162} 

\ifthreedvfinal\fi
\begin{document}

\title{3D Object Classification via Spherical Projections}

%

\author{Zhangjie Cao\\
School of Software\\
Tsinghua University, China\\
{\tt\small caozhangjie14@gmail.com}
\and 
Qixing Huang\\
Department of Computer Science\\
University of Texas at Austin, USA\\
{\tt\small huangqx@cs.utexas.edu}
\and
Karthik Ramani\\
School of Mechanical Engineering\\
Purdue University, USA\\
{\tt\small ramani@purdue.edu}
}

\maketitle

\begin{abstract}
In this paper, we introduce a new method for classifying 3D objects. Our main idea is to project a 3D object onto a spherical domain centered around its barycenter and develop neural network to classify the spherical projection. We introduce two complementary projections. The first captures depth variations of a 3D object, and the second captures contour-information viewed from different angles. Spherical projections combine key advantages of two main-stream 3D classification methods: image-based and 3D-based. Specifically, spherical projections are locally planar, allowing us to use massive image datasets (e.g, ImageNet) for pre-training. Also spherical projections are similar to voxel-based methods, as they encode complete information of a 3D object in a single neural network capturing dependencies across different views. Our novel network design can fully utilize these advantages. Experimental results on ModelNet40 and ShapeNetCore show that our method is superior to prior methods.
\end{abstract}

\section{Introduction}

We perceive our physical world via different modalities (e.g., audio/text/images/videos/3D models). Compared to other modalities, 3D models provide the most accurate encoding of physical objects. Developing algorithms to understand and process 3D geometry is vital to automatic understanding of our physical environment. Algorithms for 3D data analysis and processing were predominantly focused on hand-crafted features or shadow networks, due to limited training data we had. However, the status started to change as we have witnessed the significant growth of 3D data during the past few years (e.g., Warehouse 3D~\footnote{https://3dwarehouse.sketchup.com/?hl=en} and Yobi3D~\footnote{https://www.yobi3d.com/}), which offers rich opportunities for developing deep learning algorithms to significantly boost the performance of 3D data understanding.

Deep neural networks usually take vectorized data as input. This means how to encode input objects in vectorized forms is crucial to the resulting performance. While this problem is trivial for other modalities, e.g., audio signals, images and videos are already in vectorized forms, it becomes far more complicated for 3D objects, which are intrinsically 2D objects embedded in 3D ambient spaces. Existing deep learning algorithms fall into two categories, 3D-based and image-based. 3D-based methods typically encode a given 3D object using an occupancy grid. Yet due to memory and computational restrictions, the resolution of these occupancy grids remains low (e.g., 40x40x40 at best among most existing methods). Such a low resolution prohibits the usage of local geometric details for geometric understanding. In contrast to 3D-based approaches, image-based methods analyze and process 3D models via their 2D projections. Image-based techniques have the advantage that one can utilize significantly higher resolution for analyzing projected images. Moreover, it is possible to utilize large-scale training data (e.g., ImageNet) for pre-training. It turns out that with similar network complexity, image-based techniques appear to be superior to 3D-based techniques. Yet, there are significant restrictions of existing image-based techniques. For example, we need to determine the viewpoints for projection. Moreover, the projected images process discontinuities across image boundaries. Finally, existing techniques do not capture dependencies across different views, e.g., the correlation between the front and the back of an object, for geometric understanding.

In this paper, we introduce a novel projection method, which possesses the advantages of existing image-based techniques yet addresses the issues described above. The basic idea is to project 3D objects on a viewing sphere. On one hand, spherical domains are locally 2D, so that we can develop convolutional kernels at high resolution and utilize large-scale image data for pre-training. On the other hand, spherical domains are continuous and global, allowing us to capture patterns from complete 3D objects that are usually not present in standard image-based projections. Such characteristics make spherical projection advantageous compared with standard image-based projections. To fully utilize large-scale image training data, we present two spherical projection methods, one captures the depth variance in shapes from different view points, and the other captures the contour information of shapes from different view-points. These two projections utilize the textural and object boundary information that is captured by pre-trained neural network models from ImageNet. 

We introduce two principled ways to utilize these spherical projections for the task of 3D object classification. The guiding principle is to perform convolutions on cylindrical patches, which admit standard 2D neural network operators and thus allow us to use pre-trained neural networks. We show how to sample cylindrical patches that minimize the number of cylindrical patches, and yet are sufficient to capture rich cross-view dependencies. 

We have evaluated the proposed classification network on ModelNet40~\cite{DBLP:conf/cvpr/WuSKYZTX15} and ShapeNetCore~\cite{DBLP:journals/corr/ChangFGHHLSSSSX15}. Experimental results show that the proposed approach leads to results that are superior to or competing against state-of-the-art methods. 

\section{Related Works}

3D object classification has been studied extensively in the literature. While early works focus on leveraging handcrafted features for classification~\cite{Osada:2002:SD,2004:PSB,Knopp:2010:HTS,Huang:2013:FSL,Pu:2006:VSB,Iyer:2005:TSS}, recent methods seek to leverage the power of deep neural networks~\cite{DBLP:conf/cvpr/WuSKYZTX15,su15mvcnn}. For simplicity, we only provide a summary of methods that use deep neural networks, as they are most relevant to the context of this paper. Existing deep learning methods for 3D object classification fall into two categories: 3D-based and image-based.

\noindent\textbf{3D-based methods} classify 3D shapes based on 3D geometric representations such as voxel occupancy grids or point clouds. In~\cite{DBLP:conf/cvpr/WuSKYZTX15}, Wu et al. propose to represent a 3D shape as a probability distribution of binary variables on a 3D voxel occupancy grid and apply Deep Belief Network for classification. Recent methods utilize the same data representation but apply 3D convolutional neural networks for classification~\cite{maturana2015voxnet,DBLP:conf/ijcnn/Garcia-GarciaGR16,DBLP:journals/corr/AlvarZB16, DBLP:conf/cvpr/QiSNDYG16,brock2016generative}. They differ from each other in terms of specifications of the training data (e.g., with front orientation or without front orientation) as well as details of network training. ORION~\cite{DBLP:journals/corr/AlvarZB16} adds an orientation estimation module to the original VoxNet~\cite{maturana2015voxnet} as another task and trains both tasks simultaneously, which boosts the performance of VoxNet. Volumetric CNN~\cite{DBLP:conf/cvpr/QiSNDYG16} proposes two approaches to improve the performance of volumetric convolutional neural networks. The first one adds a sub-volume supervision task, which simultaneously trains networks that understand object parts as well as a network that understands the whole object. The second approach exploits an anisotropic probing kernel, which serves as a projection operator from 3D objects to 2D images. The results of the 2D projections can then be classified using 2D CNNs. The difference between our method and that of Volumetric CNN lies in the representation used for integrating 2D and 3D training data.  Voxception-ResNet~\cite{brock2016generative} designs a volumetric residual network architecture. To maximize the performance, it augments the training data with multiple rotations of the input and aggregates predictions from multiple residual networks.  In addition to the representations of 3D convolution neural networks, people have proposed Beam Search~\cite{DBLP:journals/corr/XuT16} to optimize the model structure and hyper-parameters of 3D convolutional networks. The basic idea is to define primitive actions to alter the model structures and hyper-parameters locally, so as to find the best model structure and parameter setting for 3D objects represented by 3D voxel grids. 

Despite the significant advances in 3D convolution, existing techniques possess a major limitation --- the resolution of a 3D convolutional neural network is usually very coarse. Although this issue has been recently alleviated by Octree-based representations~\cite{DBLP:journals/corr/RieglerUBG17,DBLP:journals/corr/RieglerUG16,DBLP:journals/corr/TatarchenkoDB17}, the cost of 3D volumetric neural networks is still significant higher than 2D neural networks of the same resolution. 

Besides voxel-based representations, people have looked at other geometric representations such as point-based representations. In~\cite{DBLP:journals/corr/QiSMG16}, the authors propose a novel neural network architecture that operates directly on point clouds. This method leads to significant improvement in terms of running time. The major challenge of designing neural networks for point cloud data is to ensure permutation invariance of the input points. In an independent work, SetLayer~~\cite{DBLP:journals/corr/RavanbakhshSP16} concentrates on the permutation equivalent property of point cloud and introduces a set-equivariant layer for this purpose. 
\begin{figure*}[htb]
\includegraphics[width=1\textwidth]{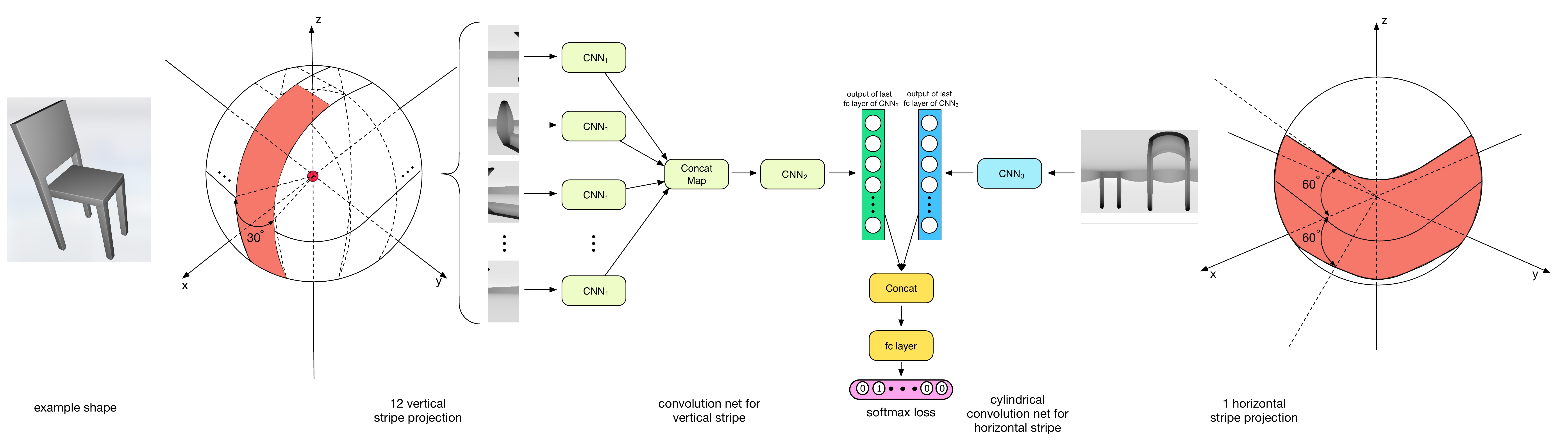}
\caption{Illustration of the depth-based projection network. The network takes a spherical projection of the an input object as input. It applies convolution operations on cylindrical strips. The output of these sub-networks are passed through a fully connected module, which captures data dependencies across different strips.}
\label{Fig:Spherical:Depth_Network}
\end{figure*}



\noindent\textbf{Image-based techniques.} The key advantage of 3D-based techniques is that the underlying 3D object can be exactly characterized by the corresponding 3D representation. On the other hand, 3D training data remains limited compared to the amount of 2D training data we have access to. In addition, such 3D representations are unable to utilize the large amount of labeled images, which can be considered as projections of the underlying 3D objects. This is the motivation of image-based 3D object classification techniques, which apply 2D convolutional neural networks to classify rendered images. In~\cite{su15mvcnn}, Su et al. propose to render 12-views for 3D shapes and classify the rendered images. The image classification network is initialized using VGG~\cite{simonyan2014very}, pretrained on ImageNet data and then fine-tuned on the ModelNet40 dataset. \cite{DBLP:conf/cvpr/JohnsLD16} provides a different way to fine-tune the network with rendered images, each of which is given a weight to measure its importance to the final prediction. \cite{DBLP:conf/eccv/SinhaBR16} proposes a way to convert 3D objects to geometry images and implicitly learn the geometry and topology of 3D objects. Despite the fact that image-based techniques can utilize pre-trained image classification networks, image projections present significant information loss, and it is not easy to capture complete relative dependencies, e.g., those that can not be projected to the same view. Perhaps the most relevant to our method is classifying panoramas~\cite{DBLP:journals/spl/ShiBZB15,3dor.20171045}, which projects a 3D object onto a cylindrical domain. Although cylindrical domain is certainly more flexible than image-domains, it still does not cover the entire object. Our experimental results reveal that a single cylindrical projection is insufficient for obtaining state-of-the-art object classification performance. 

\noindent\textbf{Hybrid methods.} Several works seek to combine 3D-based techniques and image-based techniques. In particular, FusionNet~\cite{DBLP:journals/corr/HegdeZ16} utilizes both 3D voxel data and 2D projection data by training two 3D CNNs: general 3D CNN and kernel-based 3D CNN with varying size. FusionNet also trains an image-based multi-view CNN network mentioned above. FusionNet then fuses features from these three networks so as to exploit advantages of different features.

We observe that methods based on 2D projections tend to perform better than those based on volumetric representations since they can exploit pre-trained models to address the issue of insufficient training data. Yet, image-based techniques require a large number of views. There are significant overlaps across views, exhibiting information redundancy. Thus, we propose a novel spherical projection approach, which uses a single sphere to aggregate information from different viewing directions. We also explore data dependencies across different views, which are beneficial for object classification.   


\section{Spherical Projection}

In this section, we describe the two proposed spherical projections, i.e., depth-based projection and image-based projection. The input to these two projections is a 3D model with prescribed upright orientation. However, we do not assume the front orientation is given. Such a setup is applicable to almost all internet 3D model repositories (e.g., Warehouse3D and Yobi3D).  Both spherical projections utilize a sphere centered around the barycenter of each object. The radius of this sphere is chosen as three times the diagonal of the object bounding box. Note that the radius of the sphere does not affect the depth-based projection and has minor effects on the image-based projection.

\noindent\textbf{Depth-based projection.} The depth-based projection is generated by shooting a ray from each point on the sphere to the center. Each point is recorded as the distance to the first hitting point. Otherwise, the distance is set to be zero. 
We compute depth values for vertices of a semi-regular quad-mesh whose axis aligns with the longitude and latitude, i.e., 
\begin{equation}
\begin{small}
\label{Eq:1}
\begin{aligned}
 & (\cos(\theta_i)\cos(\phi_j), \cos(\theta_i)\sin(\phi_j), \sin(\theta_i) ) \\
 &\quad \theta_i = \frac{180^\circ\cdot i}{m},  \phi_j = \frac{360^\circ\cdot j}{n}, \quad 
\begin{array}{c}
0\leq i \leq m-1 \\
0\leq j \leq n-1
\end{array}   
\end{aligned}
\end{small}
\end{equation}
Then the depth value of other points on the sphere are generated by linear interpolation. Specifically, denote $d_{ij}$ as the depth value that corresponds to $(\theta_{i}, \phi_j)$. Then given a point with spherical coordinate $(\theta, \phi)$, where $\theta_{i}\leq \theta\leq \theta_{i+1}, \phi_j \leq \phi \leq \phi_{j+1}$, its depth value is given by
\begin{align}
d = & (1-t_{ij})\big((1-s_{ij})d_{ij} + s_{ij}d_{i,j+1}\big) \nonumber \\
    & + t_{ij}\big((1-s_{ij})d_{i+1,j} + s_{ij}d_{i+1,j+1}\big).
\end{align}
This allows us to generate the depth value for every single point on the sphere. In our implementation, we further use an Octree to accelerate the ray-mesh intersection. For all of experiments, we use $n=180,m=90$, i.e., one pixel per 2 degrees along both the latitude and longitude.
We proceed to generate cylindrical strips from the depth projection described above. We first use the strip covering the following area:
\begin{equation}
\begin{small}
\label{Eq:2}
\begin{aligned}
 & (\cos(\theta_{i_h})\cos(\phi_{j_h}), \cos(\theta_{i_h})\sin(\phi_{j_h}), \sin(\theta_{i_h}) ) \\
 &\quad \theta_{i_h} = \frac{120^\circ\cdot {i_h}}{m_h}+30^\circ,  \phi_{j_h} = \frac{360^\circ\cdot {j_h}}{n_h}, \quad 
\begin{array}{c}
0\leq i_h \leq m_h-1, \\
0\leq j_h \leq n_h-1.
\end{array}   
\end{aligned}
\end{small}
\end{equation}
Since regions of high latitude suffer from severe distortion, we eliminate them by setting $\theta_{i_h}$ from $30^\circ$ to $150^\circ$. In the following, we will call this strip the latitude strip, which is fitted into the convolution layers (See Figure~\ref{Fig:Spherical:Depth_Network}). 

To utilize information form high latitude regions for classification, we also utilize a circle of vertical strips parallel to a longitude (See Figure~\ref{Fig:Spherical:Depth_Network}). There are 12 strips in total, and the angle between adjacent strips is $30^{\circ}$. The pixel coordinates on each strip indexed by $k_v$ is given by
\begin{equation}
\label{Eq:3}
\begin{small}
\begin{aligned}
 & (\cos(\delta_{k_v})\cos(\theta_{i_v})\sin(\phi_{j_v})-\sin(\delta_{k_v})\sin(\theta_{i_v}), \\ 
 &\cos(\delta_{k_v})\sin(\theta_{i_v})+\sin(\delta_{k_v})\cos(\theta_{i_v})\sin(\phi_{j_v}), \\
 &\cos(\theta_{i_v})\cos(\phi_{j_v}) ) \\
 &\quad \theta_{i_v} = \frac{360^\circ\cdot {i_v}}{l_v m_v}+90^\circ-\frac{180^\circ}{l_v},  \phi_{j_v} = \frac{180^\circ\cdot {j_v}}{n_v}, \delta_{k_v} = \frac{360^\circ\cdot {j_v}}{l_v}\\
 &\quad 0\leq i_v \leq m_v-1, 0\leq j_v \leq n_v-1, 0\leq k_v \leq l_v-1
\end{aligned}
\end{small}
\end{equation}
where $l_v$ is the number of strips.
For all experiments shown in this paper, we use $n_h=360,m_h=240$ and $n_v=180,m_v=60,l_v=12$. The total number of pixels is comparable with that used in the MVCNN~\cite{su15mvcnn}.

\begin{figure}[htb]
\includegraphics[width=0.43\textwidth]{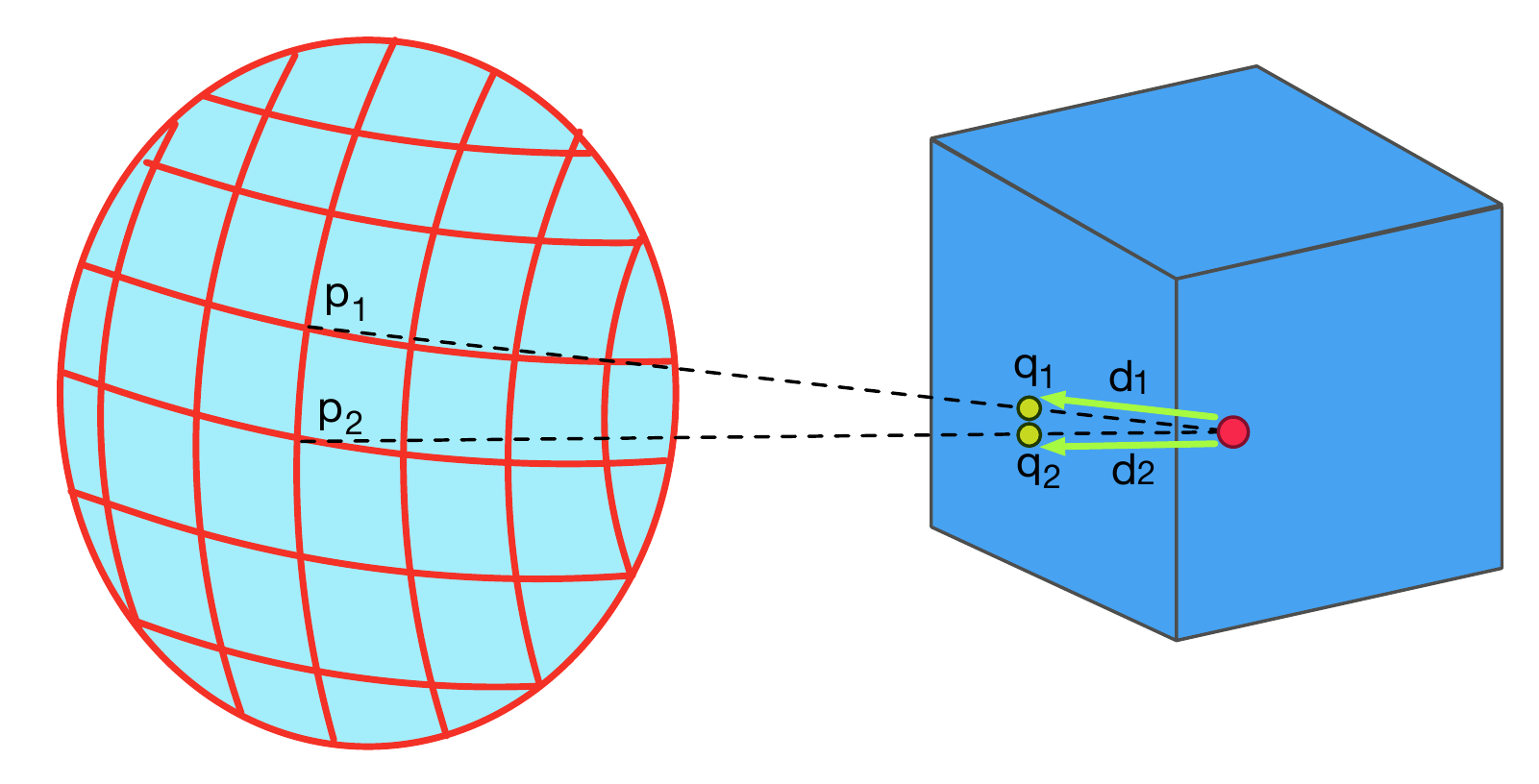}
\caption{Illustration of the depth-based projection method. We compute depth values on a rectangular grid, which is then used for interpolation.}
\label{Fig:Spherical:Depth_Projection}
\end{figure}

\begin{figure*}[htb]
\includegraphics[width=1\textwidth]{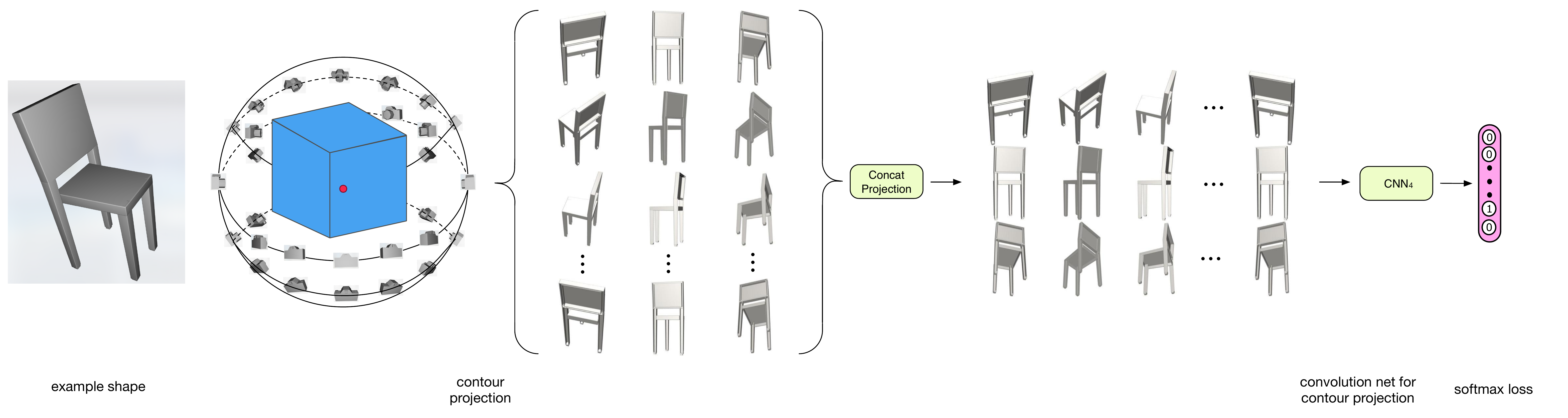}
\caption{Illustration of the contour-based projection. We utilize 36 rendered images arranged in a grid. The convolution operators are applied on discrete cylindrical strips generated from this 2D array.}
\label{Fig:Spherical:Contour_Network}
\end{figure*}

As illustrated in Figure~\ref{Fig:Spherical:Depth_Projection}, depth-based projection effectively captures geometric variations. Moreover, for a wide range of objects (i.e., the ray defined by every shape point and the sphere center reaches the sphere without occlusion), the original shapes can be directly reconstructed from the depth-based projection. Such objects include convex objects and many other box-like objects. In other words, depth-based projection is quite informative. On the downside, at the global scale, the pattern reveals in the depth-based projection seems to deviate from natural images. Moreover, the contours of objects, which provide important cues for 3D object classification, are not present in the depth-based projection. This motivates us to consider image-based projection. 

\noindent\textbf{Image-based projection.} As shown in Figure~\ref{Fig:Spherical:Contour_Network}, image-based projection shoots a 3x12 grid of images of the input object from 36 view points in total. The locations of the cameras are given by setting $m = 12, n = 3$ in (\ref{Eq:1}), i.e., $\phi_j = 0^{\circ}, 30^{\circ},\cdots, 330^{\circ}, \theta_i = -60^{\circ}, 0^{\circ}, 60^{\circ}$. At each camera location, the up-right orientation of the camera always points to the north-pole. The viewing angle of each image is $45^{\circ}$ so that the projected images barely overlap. The resolution of each image is 224x224. In our experiments, we have varied the value of $m$ and found that $m=12$ provides a good trade-off between minimizing the number of views and ensuring that the resulting projections are approximately invariant to rotating the input object. 

Note that our image-based projection does not generate a per-pixel value for each point on the sphere. Instead, we use the sphere to guide how these images are projected, enabling us to capture dependencies across different views.  

\section{Classification Network and Training}

The major motivation of the proposed network design is two-fold: 1) leveraging pre-trained convolution kernels from ImageNet, and 2) capturing dependencies that cannot be projected in the same view (e.g., front and back of an object). To this end, we propose two steps for designing the classification network.

\noindent\textbf{Network design.} Figure~\ref{Fig:Spherical:Depth_Network} illustrates the network-design for depth-based projection. The same as AlexNet, this network has a convolution module and a fully connected module. The convolution module utilizes the same set of convolution kernels as AlexNet. This allows us to use the pre-trained kernels from AlexNet. As shown in Figure~\ref{Fig:Spherical:Depth_Network}, if the strip is parallel to the latitude, the convolutions are applied in a periodical manner, so as to utilize the continuity of the data. Specifically, $CNN_3$ is a periodical convolution network which captures dependencies across different views in both convolutional and fully-connected layers. In contrast, common convolutions are applied to strips that are parallel to the longitude independently (See $CNN_1$). The fully connected layers, which are shown in $CNN_2$, capture the data dependencies across different views. To preserve the spatial relation while maintain rotation invariance, we introduce 12 fully connected layer-connections $(W_k, \bs{b}_k), 0\leq k \leq 11$ between $CNN_1$ and $CNN_2$. Let $\bs{f}_i^{l},0\leq i \leq 11$ be the feature vectors at layer $l$. We define the feature vectors at layer $l+1$ as
$$
\bs{f}_{i}^{(l+1)} = \sigma(\sum\limits_{j=1}^{12}W_{(i-j)\textup{ mod } 12}\bs{f}_j + \bs{b}_{(i-j)\textup{ mod } 12}).
$$
Note that the initial network weights $(W_0, \bs{b}_0)$ are set to be the AlexNet weights. The other weights are initialized as zero, i.e., $W_{i} = 0, \bs{i} = 0, 1\leq i \leq 11$. In other words, the initial weights apply fully connected operations on the feature vector attached to each image in isolation, while the cross links force the network to learn dependencies across different views.

Figure~\ref{Fig:Spherical:Contour_Network} shows the network design for contour-based projections. 36 Cameras are uniformly distributed along three latitudes ($60^\circ$, $90^\circ$, $120^\circ$) of the sphere. We concatenate all the rendered images in their spatial order on the sphere. We then feed the entire image to $CNN_4$, i.e., the convolutional neural network for the contour-based projection.

As described in the previous Section, the resolutions of the depth-based projections are 240x360 and 360x60x12 for strips along the longitude and the latitude, respectively. In addition, the resolution of the contour-based projection is 224x224x36. Note that although we utilize more pixels, the number of parameters in the network remains relatively small, as we share network parameters across different strips. 

\noindent\textbf{Training.} We train the entire depth network at three stages. We first train the convolutional layers and the direct connection layers, i.e., $(W_0, \bs{b}_0)$. If pre-training is allowed, the training at this stage starts from the pre-trained weights of AlexNet. We then train the convolutional network for the latitude strip. After this step, we have two pre-trained models for longitude and latitude strips. Finally, we train the entire network together with weights copied from these two pre-trained models except the final classifying layer. Note that for the contour-based network, we directly train from scratch if AlexNet parameters are not provided. If AlexNet parameters are provided, we train with all the parameters except the last classifying layer. 

We use the Caffe Framework for all of our experiments. For layers which are trained from scratch, we set its learning rate to be 10 times that of the other layers. We used mini-batch stochastic gradient descent (SGD) with 0.9 momentum and the learning rate annealing strategy implemented in Caffe. The learning rate is is cross-validated by grid search started from $10^{-5}$ and ended at $10^{-2}$, where the multiplicative step-size is ${10}^{\frac{1}{2}}$. We fix the mini-batch size as $32$ and set the weight decay as $0.0005$.
\section{Experimental Evaluation}

\subsection{Experimental Setup}

\noindent\textbf{Datasets and evaluation propocol.} We evaluate the proposed approaches on two Benchmark datasets ModelNet40 and ShapeNetCore. They both collect models from Warehouse3D but with different model classes.

\noindent\textbf{ModelNet40}~\cite{DBLP:conf/cvpr/WuSKYZTX15} contains 12311 shapes across 40 categories ranging from airplane to xbox. We use the default training-testing split (c.f. ~\cite{maturana2015voxnet, DBLP:journals/corr/AlvarZB16, DBLP:conf/cvpr/QiSNDYG16}) that leads to 9843 models in the training set and 2468 models in the testing set.

\noindent\textbf{ShapeNetCore}~\cite{DBLP:journals/corr/ChangFGHHLSSSSX15} contains 51300 shapes across 55 categories. We use the default training set (36148 models) for training and default validation set (5615 models) for testing. Note that the size of ShapeNetCore is bigger than that of ModelNet40. Distributions of categories are also different, e.g., ShapeNetCore contains less furniture categories.
 
\noindent\textbf{Baseline methods.} Since our method does not utilize the front orientation, for baseline comparison we only consider algorithms that do not utilize such information as well. In addition, we also report performance of state-of-the-art methods on ModelNet40.

\begin{table*}[!htbp]
\addtolength{\tabcolsep}{0pt}
\centering
\caption{Accuracy on ModelNet40 and ShapeNetCore of our approaches and the various baseline methods. We report the performance on these two entire datasets and subsets as well as two curated subsets.}
\vspace{10pt}
\begin{small}
\begin{tabular}{|c|c|c|c|c|}
\Xhline{1.0pt}
\centering Method & ModelNet40 & ShapeNetCore & ModelNet40-SubI & ShapeNetCore-SubI\\
\hline
3D Shapenets~\cite{DBLP:conf/cvpr/WuSKYZTX15} & 85.9 & na & 83.33 &na\\
Voxnet~\cite{maturana2015voxnet} & 87.8 & na & 85.99 &na\\
FusionNet~\cite{DBLP:journals/corr/HegdeZ16} & 90.80 & na& 89.54 &na \\
Volumetric CNN~\cite{DBLP:conf/cvpr/QiSNDYG16} & 89.9 & na & 88.65 &na\\
MVCNN~\cite{su15mvcnn} & 92.31 & 88.93 & 91.22 & 88.64 \\
MVCNN-MultiRes~\cite{DBLP:conf/cvpr/QiSNDYG16} & 93.8 & 90.01 &92.60 & 90.00\\
OctNet~\cite{DBLP:journals/corr/RieglerUG16} & 87.83 & 88.03 & 86.45 & 87.85 \\
\hline
depth-base pattern & 91.36 & 89.45 & 90.25 & 89.13\\
contour-based pattern & 93.31 & 90.49 & 92.20 & 90.80 \\
overall pattern & 94.24 & 91.00 & 93.09 & 91.22\\
\Xhline{1.0pt}
\end{tabular}
\end{small}
\label{table:overall_results}
\end{table*}
The baseline algorithms we choose include MVCNN~\cite{su15mvcnn}, MVCNN-MultiRes~\cite{DBLP:conf/cvpr/QiSNDYG16}, 3D ShapeNets~\cite{DBLP:conf/cvpr/WuSKYZTX15}, Voxnet~\cite{maturana2015voxnet}, FusionNet~\cite{DBLP:journals/corr/HegdeZ16}, Volumetric CNN~\cite{DBLP:conf/cvpr/QiSNDYG16} and OctNet~\cite{DBLP:journals/corr/RieglerUG16}. In the following, we briefly summarize the characteristics of these methods. MVCNN classifies a given model by fusing the feature layers of rendered images with a max-pooling layer. MVCNN-MultiRes improves MCVNN by exploiting rendered images from multiple resolutions. 3D ShapeNets is the first deep learning method on 3D shape data which is built on a Deep Belief Network. Voxnet leverages a 3D convolution neural network for shape classification. FusionNet fuses features extracted from 3D voxel data and 2D projection data by different networks. Finally, Volumetric CNN modifies Voxnet by adding subvolume supervision task and anisotropic probing kernel convolution. OctNet transforms 3D objects into Octree-based representations and design a special network to classify these representations. All of these methods use the upright orientation but do not use the front orientation.

\subsection{Classification Results}
Table~\ref{table:overall_results} collects the overall classification accuracy of different methods on ModelNet40 and ShapeNetCore. As we can see, the proposed depth-based projection method is superior to most existing 3D-based methods. This demonstrates the power of incorporating massive image training datasets. Compared to most other image-based techniques, our contour-driven projection method exhibits the top performance, showing the advantage of generating projections on the spherical domain. Although MVCNN-MultiRes outperforms contour-based projection, it needs to render images of different resolutions, which is much slower than our contour projection. When combining depth-based projection and contour-based projection, our overall method performs better than MVCNN-MultiRes and achieves the highest accuracy, which also demonstrates that our two projections are complementary.

When comparing the performance of various algorithms on ModelNet40 and ShapeNetCore, we find that the performance on ShapeNetCore is lower, which is expected since repositories in ShapeNetCore exhibit bigger variance. Moreover, as ShapeNetCore is bigger than ModelNet40, the gap between depth-based projection and view-based project is bigger, since the effects of pre-training may be reduced when the size of the 3D data increases. In the following, we provide more detailed analysis of the results.

\subsection{Analysis of Results}

\begin{table*}[!htbp]
\addtolength{\tabcolsep}{-0pt}
    \centering
    \caption{Accuracy Before and After Pre-training on ModelNet40}
    \vspace{10pt}
    \label{table:pre_after_modelnet}
    \begin{small}
    \begin{tabular}{|c|c|c|c|c|c|c|c|c|}
        \Xhline{1.0pt}
        \multirow{2}{30pt}{\centering Method} & \multicolumn{2}{c|}{Before Pre-training} & \multicolumn{2}{c|}{After Pre-training} \\
        \cline{2-5}
        & Accuracy (class) & Accuracy (instance) & Accuracy (class) & Accuracy (instance) \\
        \hline
        MVCNN & 82.15 & 87.15 & 90.35 & 92.31  \\
        MVCNN-MultiRes & 88.12 & 91.20 & 91.40 & 93.80 \\
        \hline
        depth-base pattern & 80.44 & 86.09 & 87.32 & 91.36 \\
        contour-based pattern & 88.33 & 91.48 & 91.16 & 93.31 \\
        overall pattern & 88.53 & 91.77 & 91.56 & 94.24 \\
        \Xhline{1.0pt}
    \end{tabular}
    \end{small}
\end{table*}

\begin{table*}[!htbp]
    \addtolength{\tabcolsep}{-0pt}
    \centering
    \caption{Accuracy Before and After Pre-training on ShapeNetCore}
    \vspace{10pt}
    \label{table:pre_after_shapenet}
    \begin{small}
    \begin{tabular}{|c|c|c|c|c|c|c|c|c|}
        \Xhline{1.0pt}
        \multirow{2}{30pt}{\centering Method} & \multicolumn{2}{c|}{Before Pre-training} & \multicolumn{2}{c|}{After Pre-training} \\
        \cline{2-5}
        & Accuracy (class) & Accuracy (instance) & Accuracy (class) & Accuracy (instance) \\
        \hline
        MVCNN & 67.84 & 84.55 & 78.79 & 88.93 \\
        MVCNN-MultiRes & 75.34 & 88.4 & 79.01 & 90.01 \\
        \hline
        depth-base pattern & 70.55 & 85.15 & 78.84 & 89.45 \\
        contour-based pattern & 74.52 & 88.54 & 79.38 & 90.49 \\
        overall pattern & 75.60 & 88.87 & 80.38 & 91.00 \\
        \Xhline{1.0pt}
    \end{tabular}
    \end{small}
\end{table*}

\noindent\textbf{The effects of pre-training.} As illustrated in Table~\ref{table:pre_after_modelnet} and Table~\ref{table:pre_after_shapenet}, all 2D-based techniques benefit from ImageNet pre-training which justifies the fact that the size of both ShapeNetCore and ModelNet40 are relative small to train high-quality classification networks, and images from ImageNet contain rich information that can be used to differentiate rendered images. When comparing ShapeNetCore with ModelNet40, the effects of pre-training on ShapeNetCore is more salient than that on ModelNet40. An explanation is that ShapeNetCore exhibits higher diversity, so that ImageNet features help more. Another factor is that the distribution of ShapeNetCore categories are closer to corresponding categories in ImageNet than ModelNet40 categories.

Quite surprisingly, the improvement of pre-training on depth-based projection is as strong as that on contour-driven projection. This means that pre-trained ImageNet models contain rich interior edge information as well (e.g., changes of texture information in the presence of depth continuities), which is beneficial for classifying depth-based projections.


\begin{table*}[!htbp]
    \addtolength{\tabcolsep}{-0pt}
    \centering
    \caption{Accuracy of Each Class For Different Projection on ModelNet40}
    \vspace{10pt}
    \label{table:class_wise}
    \begin{scriptsize}
    \vspace{0pt}
    \begin{tabular}{|c|c|c|c|c|c|c|c|}
        \Xhline{1.0pt}
        Class Name & Depth-Based & Contour-Based & MVCNN & Class Name & Depth-Based & Contour-Based & MVCNN \\
        \hline
        bowl & 100.00 & 95.00 & 85.00 & stool & 75.00 & 75.00 & 75.00\\
        bookshelf & 99.00 & 99.00 & 94.00 & tent & 95.00 & 95.00 & 95.00 \\
        cone & 100.00 & 100.00 & 95.00 & toilet & 100.00 & 100.00 & 100.00 \\
        table & 89.00 & 84.00 & 84.00 & xbox & 80.00 & 80.00 & 80.00 \\
        vase & 82.00 & 85.00 & 77.00 & car & 99.00 & 100.00 & 100.00 \\
        tv\_stand & 85.00 & 90.00 & 81.00 & guitar & 98.00 & 100.00 & 99.00 \\
        dresser & 89.53 & 89.53 & 86.05 & monitor & 97.00 & 99.00 & 99.00 \\
        bottle & 98.00 & 96.00 & 96.00 & plant & 85.71 & 89.80 & 87.76 \\
        sofa & 98.00 & 99.00 & 97.00 & range\_hood & 93.00 & 97.00 & 96.00 \\
        airplane & 100.00 & 100.00 & 100.00 & night\_stand & 75.58 & 86.05 & 80.23 \\
        bathtub & 94.00 & 96.00 & 94.00 & sink & 85.00 & 85.00 & 90.00 \\
        bed & 100.00 & 100.00 & 100.00 & piano & 91.00 & 96.00 & 97.00 \\
        bench & 80.00 & 80.00 & 80.00 & mantel & 93.00 & 97.00 & 100.00 \\
        chair & 98.00 & 99.00 & 98.00 & curtain & 85.00 & 95.00 & 95.00 \\
        desk & 86.05 & 87.21 & 86.05 & lamp & 75.00 & 80.00 & 85.00 \\
        door & 100.00 & 100.00 & 100.00 & cup & 55.00 & 80.00 & 70.00 \\
        glass\_box & 97.00 & 97.00 & 97.00 & flower\_pot & 15.00 & 15.00 & 30.00 \\
        keyboard & 100.00 & 100.00 & 100.00 & wardrobe & 65.00 & 90.00 & 90.00 \\
        laptop & 100.00 & 100.00 & 100.00 & radio & 65.00 & 95.00 & 95.00 \\
        person & 100.00 & 95.00 & 100.00 & stairs & 70.00 & 100.00 & 100.00 \\

        \Xhline{1.0pt}
    \end{tabular}
    \end{scriptsize}
\end{table*}

\noindent\textbf{Depth-based versus contour-based.} The overall performance of depth-based projection is slightly below that of contour-driven projection. This is expected because object contours provide strong cues for classification.

To further compare the effectiveness of different methods on a particular type of shapes, we selected the classes belonging to furniture from both datasets as two curated subsets, which are bathtub, bed, bookshelf, chair, curtain, desk, door, dresser, lamp, mantel, night\_stand, range\_hood, sink, sofa, stool, table, toilet, tv\_stand, wardrobe in ModelNet40 and bathtub, bed, bookshelf, cabinet, chair, clock, dishwasher, lamp, loudspeaker, sofa, table, washer in ShapeNetCore respectively. We tested our model on these subsets, and results are included in  Table~\ref{table:class_wise}.

It is clear that the winning categories of each method are drastically different. As indicated in Table~\ref{table:class_wise}, depth-based projection is advantageous on categories such as bowl, table, and bottle, which possess strong interior depth patterns. In contrast, contour-based projection is superior to depth-based projection on categories such as plant, guitar, and sofa, which have unique contour features.

\noindent\textbf{Comparison to MultiView-CNN.} The proposed contour-based projection method is superior to MultiView-CNN on both ModelNet40 and ShapeNetCore. The main reason is due to the fact our network captures dependencies earlier in the convolution, while MultiView-CNN only max-pools features extracted from the convolution layers. This performance gap indicates that dependencies across different views and at different scales are important cues for shape classification.

\noindent\textbf{Comparison to voxel-based techniques.} Our approach is also superior to most voxel-based classification methods, indicating the importance of leveraging image training data. The only exception is the recent work of Voxel-ResNet~\cite{DBLP:journals/corr/XuT16}. However, that work assumes that the front-orientation of each shape is given. In addition, its performance highly relies on training an ensemble network. In contrast, the accuracy of each individual network of~\cite{DBLP:journals/corr/XuT16}  is upper bounded by $90.1\%$.

\noindent\textbf{Comparison to panorama-based techniques.} A building block of our technique is based on classifying cylindrical strips of spherical projections. This is relevant to some recent works of classifying panoramas of 3D objects~\cite{DBLP:journals/spl/ShiBZB15,3dor.20171045}. However, the major difference in our approach is that we use multiple strips to capture the correlations of the spherical projection from multiple strips. In addition, the network design utilizes a pre-trained model from ImageNet. As indicated by our experiments, on ModelNet40 our approach leads to $4.5\%$ and $3.4\%$ improvements in terms of accuracy from using a single strip and that of~\cite{3dor.20171045}, respectively.

\begin{table}[!htbp]
    \addtolength{\tabcolsep}{-0pt}
    \centering
    \caption{Accuracy w.r.t Number of Views for Depth and Contour Pattern on ModelNet40 and ShapeNetCore}
    \vspace{10pt}
    \label{table:views}
    \begin{scriptsize}
    \begin{tabular}{|c|c|c|c|c|c|c|c|}
        \Xhline{1.0pt}
        Pattern & Number of Views & ModelNet40 & ShapeNetCore \\
        \hline
        \multirow{3}{40pt}{depth-based} & 6 & 90.87 & 88.95 \\
        & 24 & 91.02 & 89.23 \\
        & 12 & 91.36 & 89.45 \\
        \hline
        \multirow{3}{40pt}{contour-based} & 6  & 92.26 & 89.23 \\
        & 24 & 93.12 & 90.36 \\
        & 12 & 93.31 & 90.49 \\
        \Xhline{1.0pt}
    \end{tabular}
    \end{scriptsize}
\end{table}

\noindent\textbf{Varying the resolutions of the projections.} We have also tested the performance of our network by varying resolutions of the projections. For depth-based projection, we increase the resolution of the grid from $30^{\circ}$ to $15^{\circ}$ and $60^{\circ}$, we found that the classification accuracy drops by $0.2\%$ and $0.5\%$, respectively. We thus used $30^{\circ}$ for efficiency concerns. For contour-based projection, we have changed resolution of the grid pattern to $3\times24$ and $3\times6$, we found that the improvements in classification accuracy improves by less than $0.2\%$ from $3\times12$ grids to $3\times24$ grids. On the other hand, the improvement from using $3\times6$ grids to $3\times12$ grids is about $1.0\%$ on average, which is expected since using a $3\times6$ grid is insufficient for handling rotation invariance.


\begin{figure}[htb]
\caption{Accuracy w.r.t Elevation degree of the strip parallel to the latitude}
\vspace{5pt}
\includegraphics[width=0.47\textwidth]{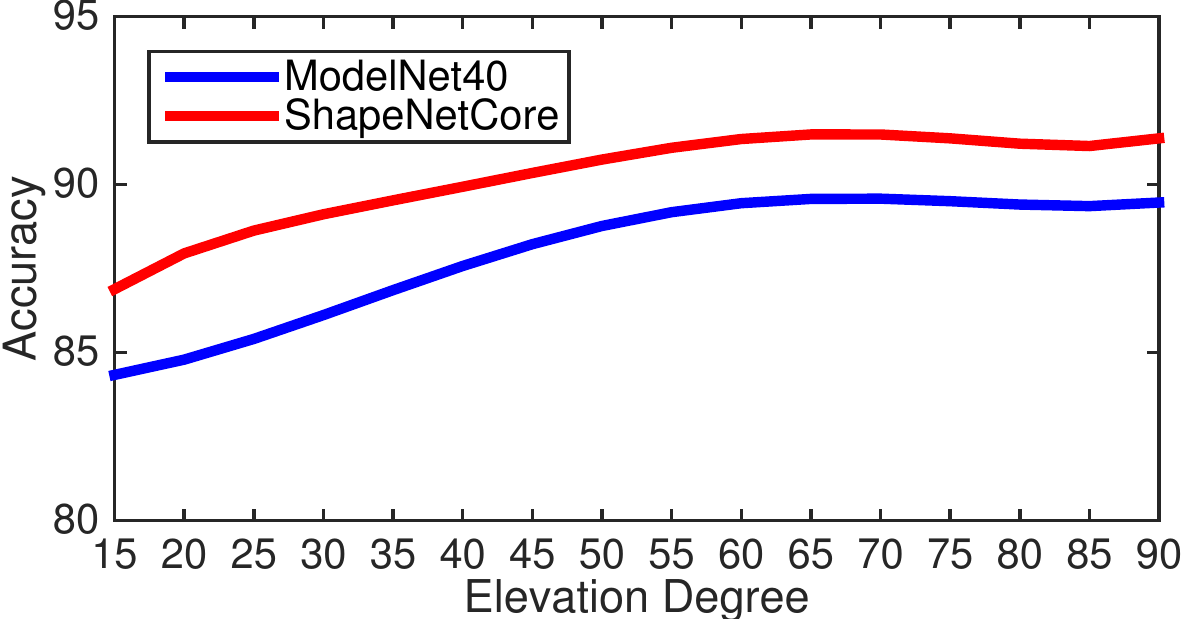}
\label{Fig:elevation}
\end{figure}

\noindent\textbf{Varying the elevation degree horizontal strip.} We also tried varying the elevation degree of the horizontal strips, which is $60^\circ$ in Figure~\ref{Fig:Spherical:Depth_Network}. Experimental results in Figure~\ref{Fig:elevation} show that the performance is not that sensitive when the elevation degree varies. We can see that when the elevation degree increases larger than $60^\circ$, the performance does not apparently increase with the increasing of the elevation degree, since high altitude areas have severe distortion in the horizontal strip, where pre-trained models become ineffective. Thus, we choose elevation degree as $60^\circ$.

\noindent\textbf{Timing.} The rendering, inference, and training time are listed in Table~\ref{Table:2}, which are performed on a machine with 16 Intel Xeon E5-2637 v4 @ 3.50GHz CPUs and 1 Titan X(Pascal) GPU. As indicated in Table~\ref{Table:2}, classifying a single 3D object takes around 2 seconds. The dominant computational cost is on generating the projections. Note that both depth-based projections and contour-based projections can be accelerated using GPU. We believe the computational cost can be significantly improved by exploring such options.
\begin{table}[!htbp]
    \addtolength{\tabcolsep}{-2.5pt}
    \centering
    \caption{Running Time of All the methods on ModelNet40 and ShapeNetCore}
    \vspace{10pt}
    \label{table:running_time}
    \begin{scriptsize}
    \begin{tabular}{|c|c|c|c|c|c|c|}
        \Xhline{1.0pt}
        \multirow{2}{20pt}{Method} & \multicolumn{3}{c|}{ModelNet40} & \multicolumn{3}{c|}{ShapeNetCore} \\
        \cline{2-7}
        & Rendering & Inference & Training & Rendering & Inference & Training  \\
        \hline
        Depth & 0.92s & 0.043s & 252m & 1.06s & 0.043 & 272m\\
        Contour & 1.17s & 0.418s & 371m & 1.28s & 0.418s & 442m \\
        \Xhline{1.0pt}
    \end{tabular}
    \end{scriptsize}
\label{Table:2}
\end{table}

\section{Conclusions, Discussion and Future Work}
In this paper, we have introduced a spherical representation and developed deep neural networks to classify 3D objects. Our approach explores two ways to project 3D shapes onto a spherical domain. The first one leverages depth variation, while the other one leverages contour information. Such a representation shows advantages of 1) allowing high resolution grid representations to capture geometric details, 2) incorporating large-scale labeled images for training, and 3) capturing data dependencies across the entire object. We also described principled ways to define convolution operations on spherical domains such that the output of the neural networks is not sensitive to the front orientation of each object. Experimental results show that the proposed methods are competitive against state-of-the-art methods on both ModelNet40 and ShapeNetCore.


There are ample opportunities for future research. The methods presented in this paper still use rectangular convolutional kernels mainly due to the fact that we want to use pre-trained convolutional kernels. However, technically it will be interesting to see if one can define convolutional kernels directly on spherical domains. One potential solution is to use spherical harmonics~\cite{Kazhdan:2003:RIS:882370.882392}. In another direction, it remains interesting to consider other types of spherical projections, e.g., spherical parameterizations of geometric objects~\cite{Hormann:2008:MPT}, which is free of occlusions. We did not use such parameterizations mainly due to that models in ModelNet40 and ShapeNetCore consist a lot of disconnected components. Finally, we only consider the task of classification, it will be interesting to consider other tasks such as shape segmentation and shape synthesis. For both tasks, the standard image-based techniques require stitching predictions from different views of the objects. In contrast, the spherical projection is complete and may not suffer from this issue.

\noindent\textbf{Acknowledgement.} We would like to acknowledge support of the NSF Award IIP \#1632154. We also gratefully acknowledge the support of NVIDIA Corporation with the donation of the Titan Xp GPU used for this research. Any opinions, findings, and conclusions or recommendations expressed in this material are those of the authors and do not necessarily reflect the views of the sponsors. 

{\small
\bibliographystyle{ieee}
\bibliography{3d_convolution}
}
\appendix

\end{document}